\documentclass{article}

\PassOptionsToPackage{numbers, compress}{natbib}

\usepackage[final]{neurips_2024}

\usepackage[utf8]{inputenc} % allow utf-8 input
\usepackage[T1]{fontenc}    % use 8-bit T1 fonts
\usepackage{hyperref}       % hyperlinks
\usepackage{url}            % simple URL typesetting
\usepackage{booktabs}       % professional-quality tables
\usepackage{amsfonts}       % blackboard math symbols
\usepackage{nicefrac}       % compact symbols for 1/2, etc.
\usepackage{microtype}      % microtypography
\usepackage{xcolor}         % colors

\usepackage{graphicx}
\usepackage{amsmath}
\usepackage{float}
\usepackage{subfigure}
\usepackage{natbib}
\usepackage{caption}

\usepackage{indentfirst}
\graphicspath{{images/}}

\usepackage{enumitem}
\usepackage{multirow}
\usepackage{array}

\title{HiCo: Hierarchical Controllable Diffusion Model for Layout-to-image Generation}

\renewcommand\footnotemark{}
\author{
Bo Cheng \quad
Yuhang Ma \quad
Liebucha Wu \quad
Shanyuan Liu \\
\textbf{Ao Ma} \quad
\textbf{Xiaoyu Wu} \quad
\textbf{Dawei Leng $^\dagger$}  \quad
\textbf{Yuhui Yin} \\
\thanks{$^\dagger$Corresponding authors.}\\
\texttt{360 AI Research} \\
\texttt{\{chengbo1, mayuhang, wuliebucha, liushanyuan\}@360.cn} \\
\texttt{\{maao, wuxiaoyu1, lengdawei, yinyuhui\}@360.cn} 
}

\begin{document}

\maketitle

\begin{figure*}[!htbp]
    \vspace{-10mm}
    \centering

    \subfigure{
        \begin{minipage}[t]{0.48\textwidth}
            \centering
            \includegraphics[width=1\textwidth]{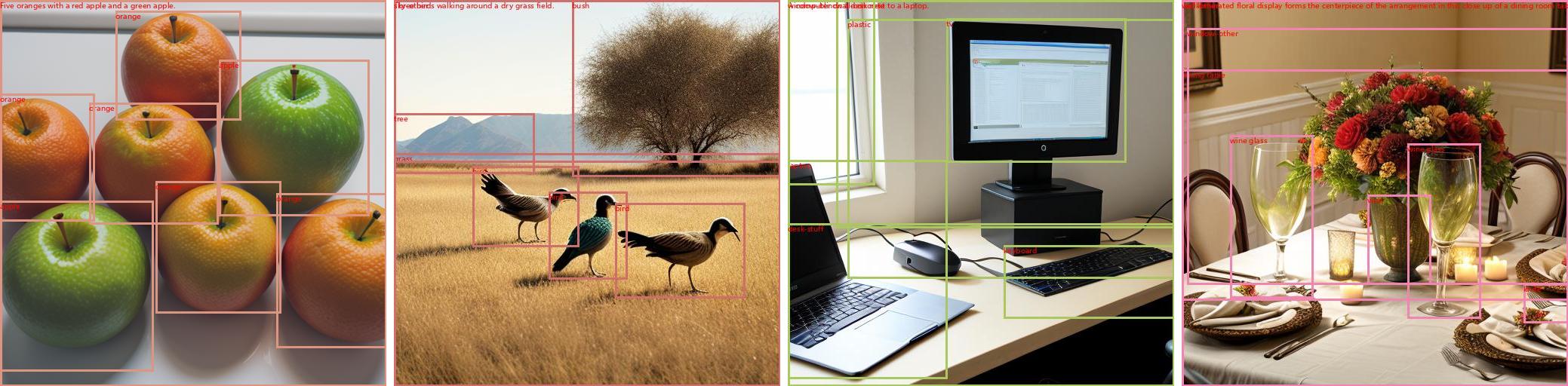}            
            \vspace{-0.6cm}
            \caption*{ (a) Layout grounded generation with closed-set short descriptions}
            \label{fig:final_combined_image_fuse_a}
        \end{minipage}
    }
	\subfigure{
        \begin{minipage}[t]{0.48\textwidth}
            \centering
            \includegraphics[width=1\textwidth]{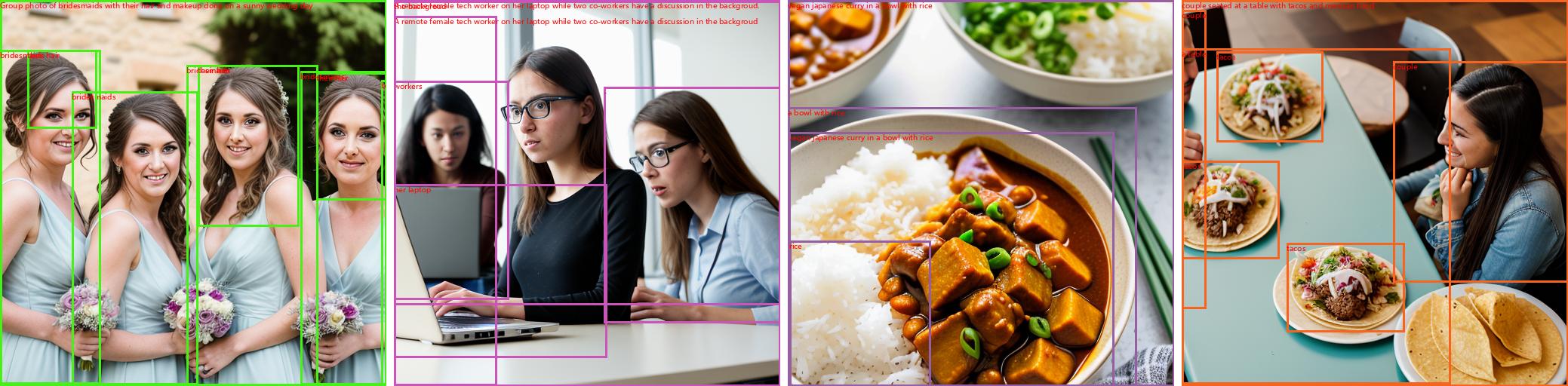}            
            \vspace{-0.6cm}
            \caption*{ (b) Layout grounded generation with open-ended fine-grained descriptions}
            \label{fig:final_combined_image_fuse_b}
        \end{minipage}
    }

	\vspace{-3mm}
    \setcounter{subfigure}{0}

    \subfigure{

        \begin{minipage}[t]{0.48\textwidth}
            \centering
            \includegraphics[width=1\textwidth]{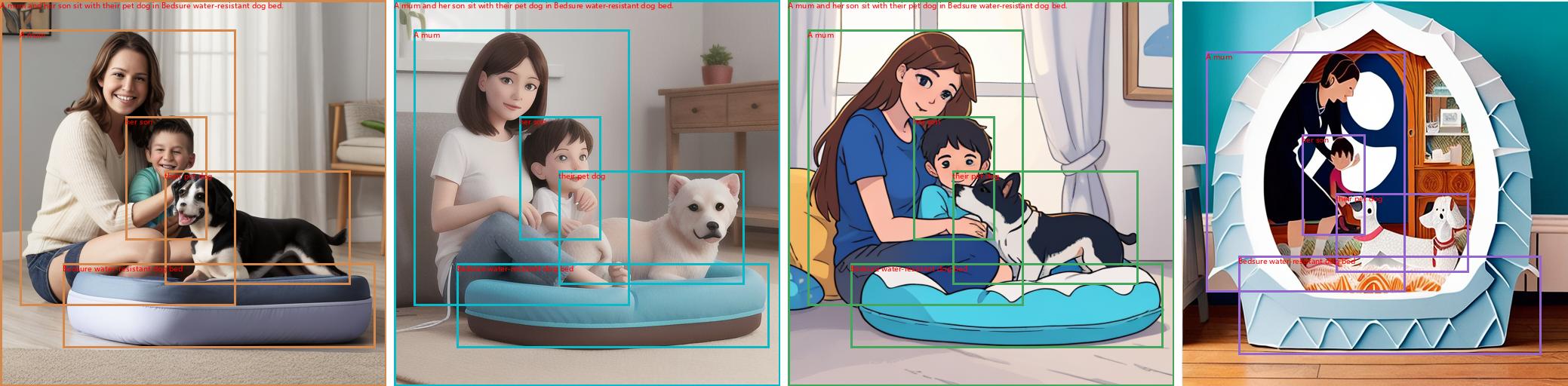}
            \phantom{Figure \thefigure}
            \vspace{-0.6cm}
            \caption*{ (c) HiCo is compatible with different SD variants}
            
            \label{fig:final_combined_image_fuse_c}
        \end{minipage}
    }
	\subfigure{
        \begin{minipage}[t]{0.48\textwidth}
            \centering
            \includegraphics[width=1\textwidth]{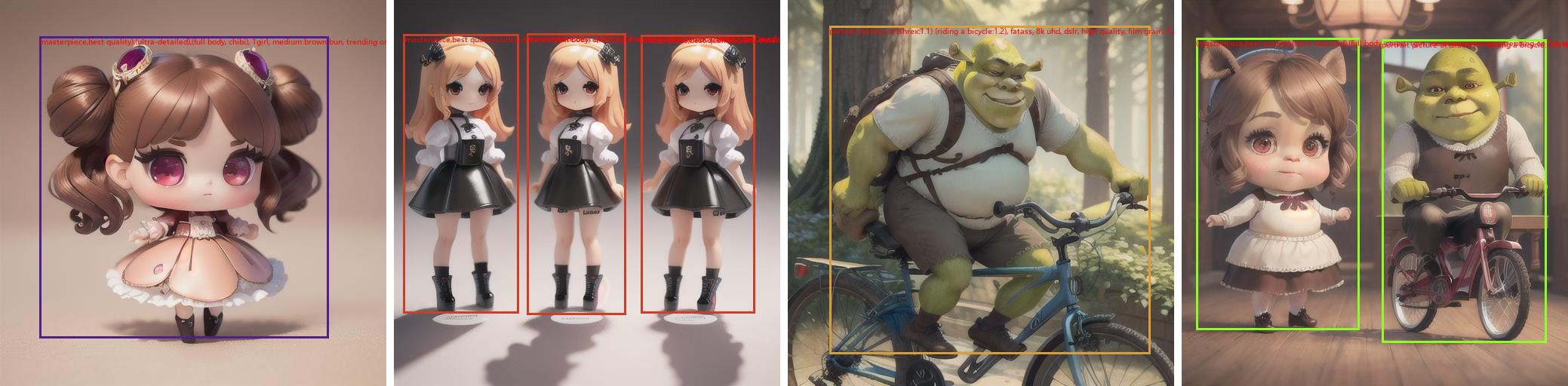}
            \phantom{Figure \thefigure}
            \vspace{-0.6cm}
            \caption*{ (d) Multi-concept generation by HiCo with multi LoRAs}
            \label{fig:final_combined_image_fuse_d}
        \end{minipage}
    }

    \vspace{-3mm}

	\setcounter{subfigure}{0}

    \subfigure{
        \begin{minipage}[t]{0.98\textwidth}
            \centering
            \includegraphics[width=1\textwidth]{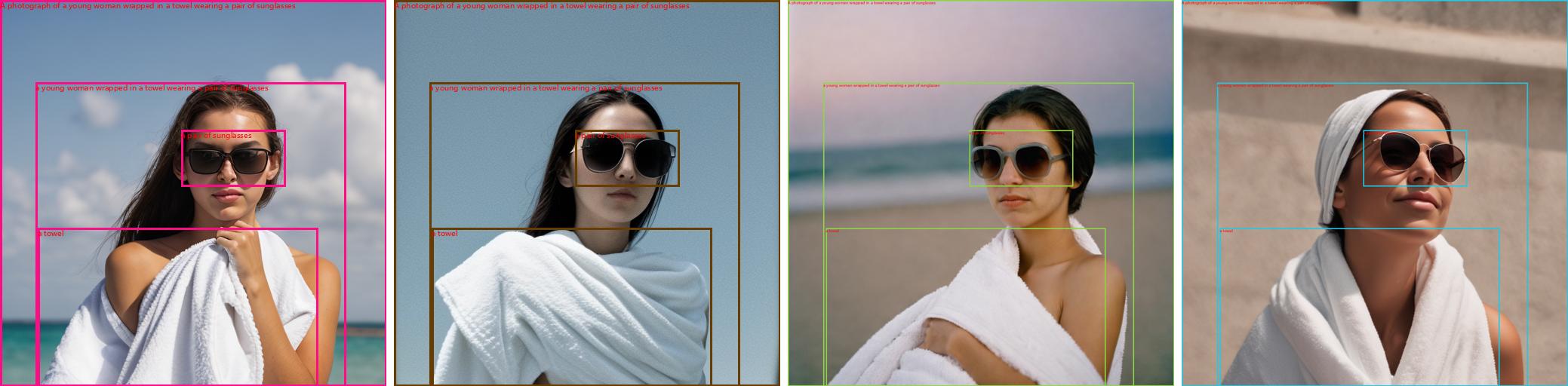}
            \phantom{Figure \thefigure}
            \vspace{-0.6cm}
            \caption*{ (e) Fast inference with HiCo-LCM / HiCo-Lightning}
            \label{fig:final_combined_image_fuse_e}
        \end{minipage}
    }
    \vspace{-3mm}
	\caption{HiCo model serves to enhance layout controllability for text-to-image generation, by integrating bounding box condition of different objects hierarchically. The proposed unique conditioning branch structure can produce more harmonious and holistic image with complex layout.}
	\vspace{-3mm}
    \label{fig:result_include1}
\end{figure*}

\begin{abstract}
   The task of layout-to-image generation involves synthesizing images based on the captions of objects and their spatial positions. Existing methods still struggle in complex layout generation, where common bad cases include object missing, inconsistent lighting, conflicting view angles, etc. To effectively address these issues, we propose a \textbf{Hi}erarchical \textbf{Co}ntrollable (HiCo) diffusion model for layout-to-image generation, featuring object seperable conditioning branch structure. Our key insight is to achieve spatial disentanglement through hierarchical modeling of layouts. We use a multi branch structure to represent hierarchy and aggregate them in fusion module. To evaluate the performance of multi-objective controllable layout generation in natural scenes, we introduce the HiCo-7K benchmark, derived from the GRIT-20M dataset and manually cleaned. \url{https://github.com/360CVGroup/HiCo_T2I}.
\end{abstract}

\section{Introduction}
Text-to-image (T2I)\citep{rombach2022high,ramesh2022hierarchical,nichol2021glide,li2023gligen} diffusion models like Stable Diffusion, GLIDE\citep{nichol2021glide}, have rapidly developed for their exceptional quality and diverse generative capabilities. However, the T2I models lack fine-grained control over visual composition and spatial layout via text prompts alone.

\begin{figure}[H]
    \vspace{-3mm}
    \centering
    \includegraphics[width=1.0\linewidth]{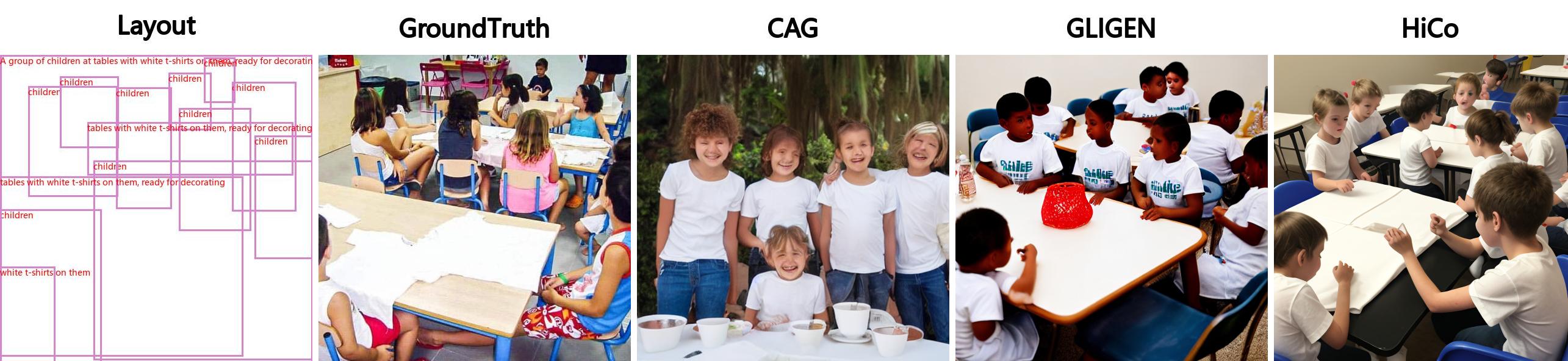}
    \caption{ The generation of CAG\citep{chen2024training}, GLIGEN\citep{li2023gligen} and HiCo in complex layouts.}
    \vspace{-5mm}
    \label{fig:HiCo_introd_1}
\end{figure}

Layout-to-image generation\citep{rombach2022high,li2023gligen,li2021image}, which aims to produce high-quality and realistic images from layout conditions. This article mainly studies the generation of layout images based on object text description and positional coordinates. Existing methods can be mainly divided into two categories: training-free methods\citep{chen2024training,zhao2023loco,xie2023boxdiff,bar2023multidiffusion} and training-based methods\citep{li2023gligen,cheng2023layoutdiffuse,zheng2023layoutdiffusion,zhou2024migc}. Training-free methods usually use cross-attention to get the ability to control position. Training-based methods typically utilize new network structures or specific attention. As shown in the Figure \ref{fig:HiCo_introd_1}, in complex scenarios, the training-free method represented by CAG\citep{chen2024training} has a serious problem of object missing. The training-based methods represented by GLIGEN\citep{li2023gligen} can alleviate the phenomenon of object missing, but the generated images often exhibit distortion.

To address the issues, we propose the Hierarchical Controllable(HiCo) diffusion model. The approach disentangle the spatial layouts by multiple branch networks. 
Specifically, the branch design of HiCo is inspired by external condition introduction methods similar to ControlNet\citep{zhang2023adding} and IP-Adapter\citep{ye2023ip}. Hierarchical layout features are separately extracted by branches of weight sharing, and refinedly aggregated with Fuse Net. The overall architecture of our approach is shown in Figure \ref{fig:HiCo}. Our method demonstrates superior performance in terms of object omissions and image quality as shown in Figure \ref{fig:HiCo_introd_1}. This is attributed to our hierarchical modeling approach, which has particular advantages in generating complex layouts.

To further validate the effectiveness of our method, we conducted experiments on both the closed-set COCO dataset and the open-ended GRIT dataset, and achieved excellent performance on both. Furthermore, our method exhibits a flexible scalability, including switching checkpoints and integrating plugins like LoRA, LCM. Refering to Figure \ref{fig:result_include1} for details.

Due to the lack of objective benchmark for multi-objective controllable layout in natural scenes, we introduce the HiCo-7K benchmark. 
HiCo-7K is uniformly sampled from GRIT-20M\citep{peng2023kosmos} dataset, revalidated for object regions using Grounding-DINO\citep{liu2023grounding}, and filtered based on semantic relevance using CLIP\citep{radford2021learning}. Furthermore, we conduct deep manual cleaning on it.

Our primary contributions are shown as following:
\begin{enumerate}[leftmargin=*]
    \item We propose the HiCo model, which achieves spatial disentanglement through hierarchical modeling of layouts. Our method can generate more desirable images in complex scenarios, and exhibit a flexible scalability.
    \item We propose a benchmark HiCo-7K, which has been revalidated and cleaned by algorithms and professionals. It can objectively evaluate the task of layout image generation in natural scenes.

    \item Our method achieves state-of-the-art performance on both the open-ended HiCo-7K dataset and the closed-set COCO-3K\citep{caesar2018coco} dataset.
\end{enumerate}

\section{Related Work}

\subsection{Diffusion Models}
Diffusion models are generative models that synthesize images from random noise by iterative image denoising. They have showed its excellent potential to generate high semantic relevance and aesthetic quality images than GAN-related models\citep{creswell2018generative}. DDGAN\citep{chen2023ddgan}, DiffusionVAE\citep{pandey2022diffusevae} study the combination of diffusion model and other generative methods. Compared to denoising diffusion probabilistic models (DDPM)\citep{ho2020denoising}, DDIM\citep{song2020denoising} and PLMS\citep{liu2022pseudo} mitigate the lengthy sampling procedure by reducing number of sampling steps. The latent diffusion model(LDM)\citep{rombach2022high} leverages VAE to encode images to latent codes with smaller resolution, saving efforts to train super-resolution models for generating high-resolution images. ControlNet\citep{zhang2023adding} and IP-Adapter\citep{ye2023ip} enabled additional image-guided conditions into frozen T2I diffusion models (e.g., sketch, segmentation masks, canny edge).

\subsection{Layout-to-Image Generation}
Layout-to-image technology seeks to generate realistic images with corresponding spatial layouts based on graphical or textual inputs that convey layout information. Early layout-to-image techniques primarily used GAN-related technologies\citep{karras2019style,sun2021learning,wang2022interactive}. Although these works achieved encouraging progress, their generation effects and application scenarios were extremely limited.

Different guiding conditions\citep{balaji2022ediff,brooks2023instructpix2pix,hertz2022prompt,epstein2023diffusion} endow diffusion models with diverse capabilities. Researchers have proposed various methods to generate layout-controllable images using object descriptions and spatial positions. They mainly design a brand new network or special attention, such as layout attention or attention redistribution. Our approach build on basic pre-trained model by introducing weight-shared multi-branch structures for enhanced local controllability.

Researches\citep{he2024llms,fu2023guiding} on the combination of large language model(LLM) and diffusion model, can enhance strong performance in instruction-following and controllable image generation and editing tasks.Unlike HiCo which requires layout and image specification directly from user input, LMD\citep{lian2023llm} and SLD\citep{wu2024self} resort to LLM for image scene description and layout arrangement via text automatically. It’s worth pointing out that HiCo can be integrated with LMD and SLD, by treating HiCo as the replacement of their train-free layout controllable image generation module.

\section{Method}

\subsection{Preliminary}
The SD model is a diffusion model that operates within the latent space, which consists of three main components: The VAE-encoder\citep{van2017neural} encodes images into the latent space, while the VAE-decoder reconstructs the latent representation into realistic images. The CLIP\citep{radford2021learning,dosovitskiy2020image} text encoder projects a sequence of tokenized texts into the sequence embedding. The UNet model is trained to predict the added Gaussian noise using LDM loss.

% \vspace{-5mm}
\begin{equation}
    L_{LDM} = E_{\epsilon(x),t,\epsilon \sim N(0,1)}\left[||\epsilon-\epsilon_{\theta}(z_t,t)||^2_2\right]
\end{equation}
where \(t\) is uniformly sampled from time steps \( \{1,\ldots,T\} \), \(z_t\) is \(t\)-step latent of the input. \(\epsilon_{\theta}\) is noise prediction model.

The conditional guided generation in SD involves incorporating text conditions, reference images, masks, and other conditions into the SD model through various techniques. Controlnet is a common method of introducing external control conditions through collateral structures, and its training goal is to predict noise at different stages with a learnable branch network, denoted as \(\epsilon_{\theta}\). Given the latent \(z_0\), the model adds noise to reach \(z_t\), where \(t\) represents the number of noise-adding steps. Here, \(c_t\) indicates the textual control condition, and \(c_f\) stands for a specific control condition. The objective function can be expressed as a function \(L_{condition}\).

% \vspace{-5mm}
\begin{equation}
    L_{condition} = E_{z_0,t,c_t,c_f,\epsilon \sim N(0,1)}\left[||\epsilon-\epsilon_{\theta}(z_t,t,c_t,c_f)||^2_2\right]
\end{equation}
% \vspace{-5mm}

\subsection{HiCo Diffusion Model}
\begin{figure}[H]
    \vspace{-5mm}
    \centering
    \includegraphics[width=0.9\linewidth]{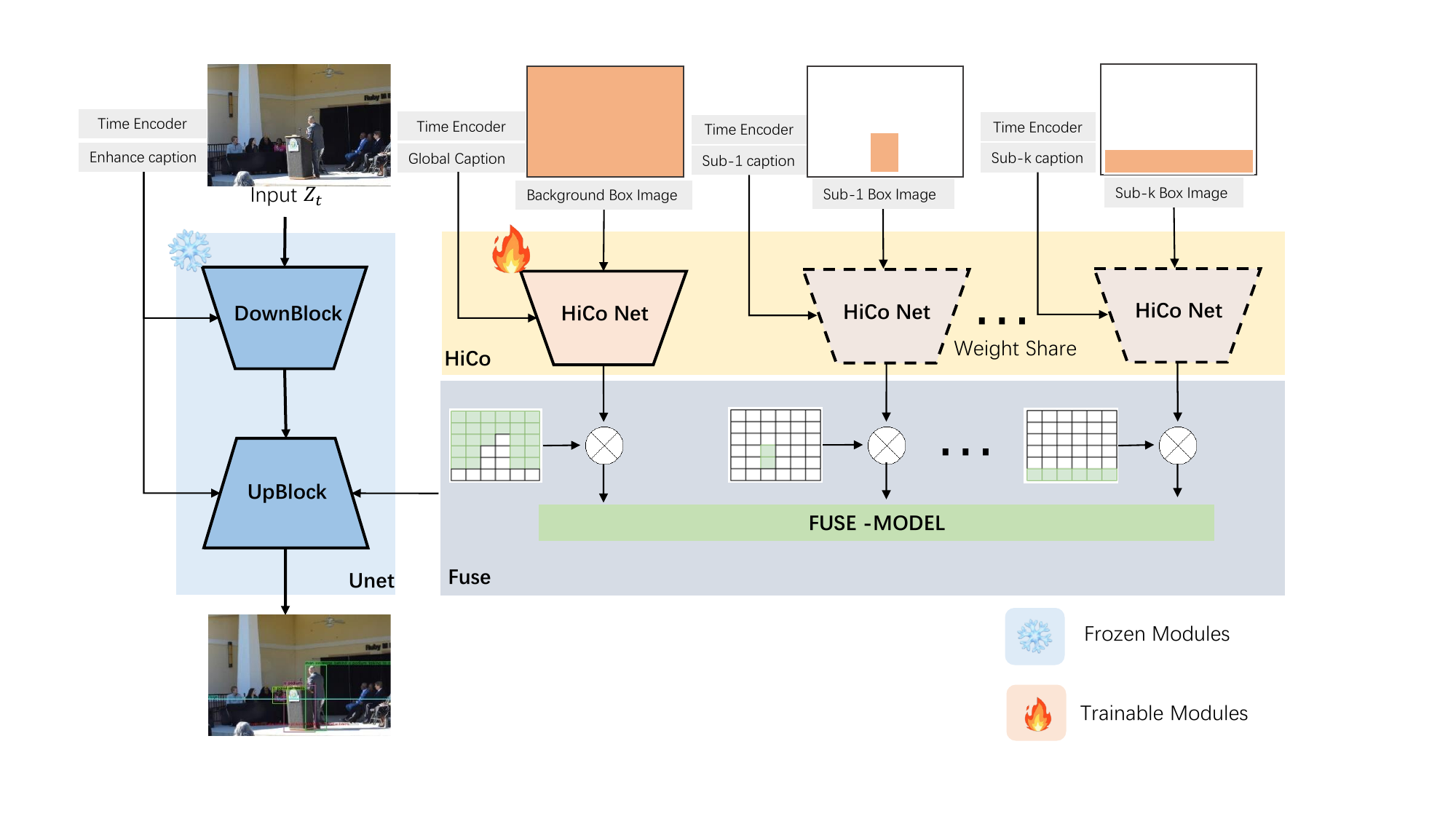}
    \vspace{-5mm}
    \caption{The overall architecture of our approach.}
    \vspace{-5mm}
    \label{fig:HiCo}
\end{figure}

We adopt a common external condition introduction method similar to ControlNet and IP-Adapter, and explore its innovative application in the design of controllable layout networks. We proposed a multi-branch HiCo Net, which independently models the background and multiple foregrounds, and hierarchically expresses the local semantics and spatial layout relationship of the image in a fine-grained manner. On the fusion of branches, we experimented with a variety of fusion methods and proposed a non-parametric Fuse Net, decouples branches through masks and achieves excellent performance. The overall network structure is shown in Figure \ref{fig:HiCo}.

\textbf{HiCo Net}.The multi-branch HiCo Net is introduced to generate the global background and foreground instances for different layout regions. The branch of HiCo Net adopts the structure of ControlNet, independently decoupling layout conditions to hierarchically model foreground objects.

\begin{equation}
    L_{HiCo} = E_{z_0,t,c^k_t,c^k_b,\epsilon \sim N(0,1),k \sim [1,K]}\left[||\epsilon-\epsilon_{\theta}(z_t,t,\mathcal{F}(c^k_t,c^k_b)||^2_2\right]
\end{equation}

Here, \(c^k_t\) represents the textual control condition of \(k\)-th sub area, and \(c^k_b\) represents the bounding box control condition of \(k\)-th sub area. \(F\) represents the Fuse Net.

We define the instruction to a layout-to-image model as a composition of sub-caption and sub-boundingbox.
\begin{equation}
    \begin{aligned}
        \text{Instruction}: &y=(c_i, b_i),i \in [1,K], with \\
        \text{caption}:     &c=[c_g,c_1,...,c_i,...,c_K] \\
        \text{boundingbox}: &b=[b_g,b_1,...,b_i,...,b_K]\\
    \end{aligned} 
\end{equation}

where \(K\) represents the number of regions, with  \(c_g\) and \(b_g\) representing the textual descriptions and positions of the background image, \(c_i\) and \(b_i\) corresponding to the descriptions and positions of the individual regions. The HiCo Net processes the different regional conditions to generate intermediate features that correspond to the textual descriptions within the predefined regions.

\textbf{Fuse Net}.The module fuses intermediate features from sub-regions and acts as the external features of the UNet model. It have different forms, including summation, averaging, mask, learnable-weights, and other methods. Based on different task types, various fusion structures can be selected. Our approach mainly decouples the content of different foreground and background regions through the mask fusion form. The detailed fusion process is defined as:

\begin{equation}
    \begin{aligned}
        \mathcal{F}(c_t,c_b)=&M_g\cdot\epsilon_{hico}(z_t,t,c_t^g,c_b^g)+\sum_{k=1}^KM_k\cdot\epsilon_{hico}(z_t,t,c_t^k,c_b^k), M_g=1-\sum_{k=1}^KM_k 
    \end{aligned}
    \label{eqn:equation_1}
\end{equation}
Here, $c_t^k$, $c_b^k$ represent the text and position control conditions of \textit{k-th} sub area, and $c_t^g$, $c_b^g$ represent the textual description and position of the background image. $M_k$ represents the \textit{k-th} object area mask information, and $M_g$ represents the background area mask information.

\begin{figure}[H]
    \vspace{-5mm}
    \centering
    \includegraphics[width=0.9\linewidth]{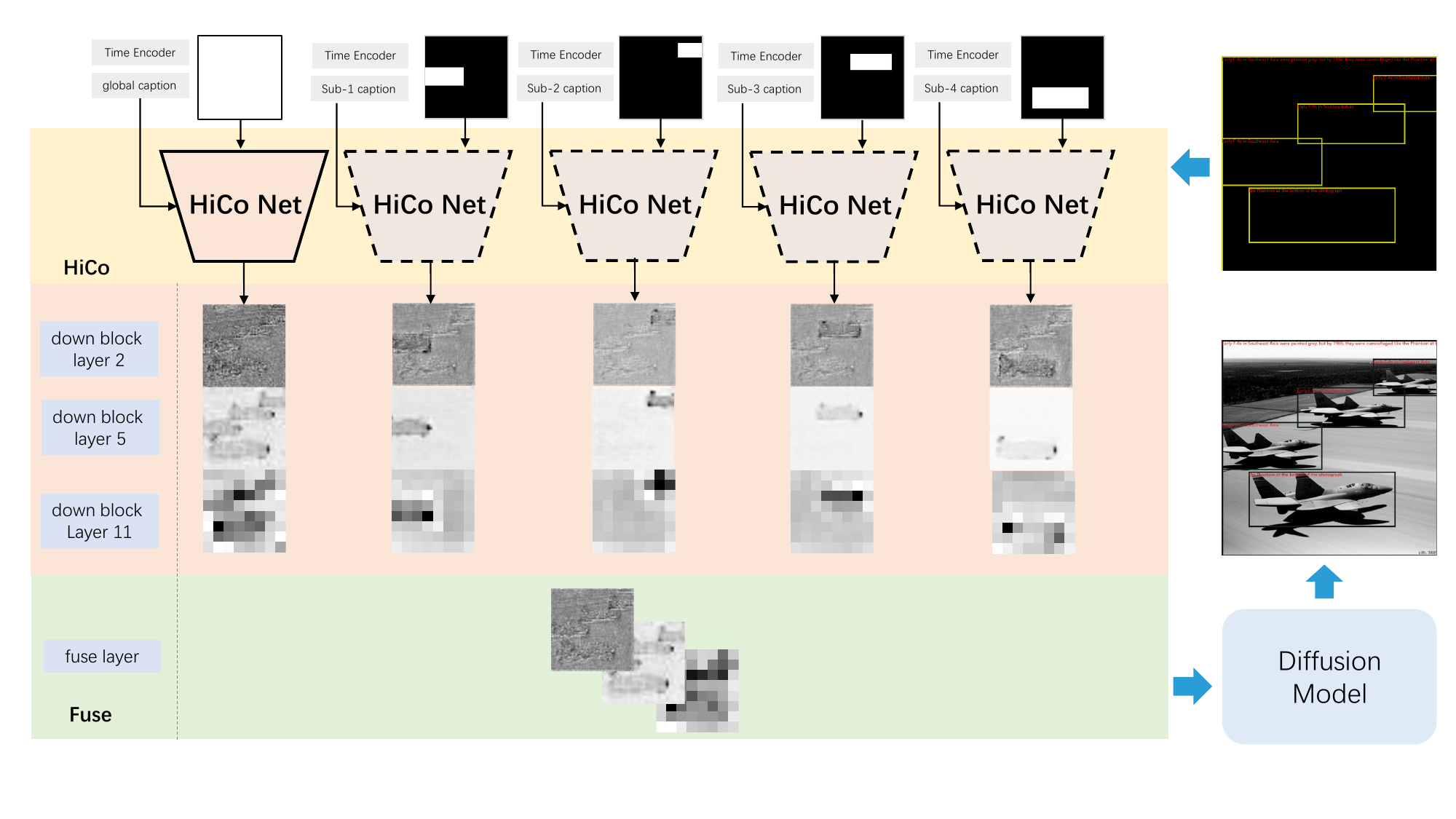}
    \caption{The visualization of the features on different layers of the HiCo branch and Fuse Net.}
    \vspace{-5mm}
    \label{fig:HiCo Vis}
\end{figure}

\begin{figure}[H]
    \vspace{-5mm}
    \centering
    \includegraphics[width=0.9\linewidth]{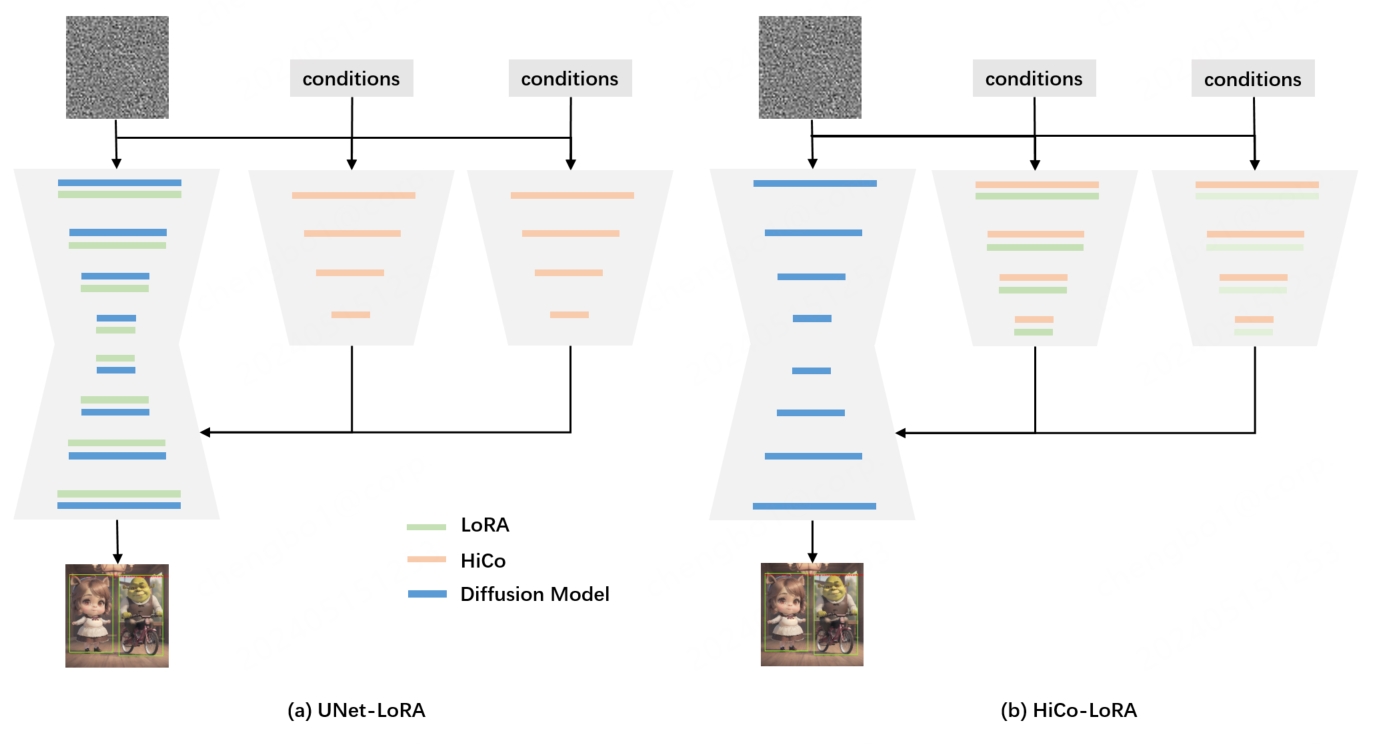}
    \caption{The model fine-tuning technique based on various positions of LoRA. (a) Adding LoRA parameters on UNet. (b) Adding LoRA parameters on HiCo.}
    \vspace{-5mm}
    \label{fig:HiCo Lora}
\end{figure}

\subsection{Hierarchical Controllable Analysis}
The introduction of external control conditions through a side branch structure is a common condition guidance approach in diffusion models. Our HiCo model employs an innovative weight-sharing mechanism across its branch structures, which adeptly generates distinctive features for both foreground objects and the background image, tailored to the specific conditions of the caption and bounding box. These features are strategically integrated during the upsampling phase, a critical step in the diffusion model's generative workflow.

Figure \ref{fig:HiCo Vis} depicts the generative process of the HiCo model for four foreground objects, showcasing the visualization of features of layer-2, layer-5, and layer-11 at the 50th denoising step during the downsampling stage. The visualization of shallow features reveal that various branches exhibit a more pronounced response to their respective layout areas. The intermediate features indicate further refinement of object positions, contours, and semantics generated by different branches. Furthermore, the deep features underscore the model's adept handling of regional information, essential for the controlled layout of the image. The fusion process of the HiCo branches employs a mask technique. However, it is crucial to elucidate how this fusion contributes to the overall semantic coherence of the spatial layout.

\subsection{Expansion Capability} \label{Expansion Capability}

Low-Rank Adaptation (LoRA)\citep{ryu2023low} stands out as an efficient fine-tuning technique. Our HiCo model exhibits excellent compatibility with rapid generation plugins, whether it's utilizing LCM-LoRA\citep{luo2023lcm}for quickly generating 512$\times$512 resolution images on HiCo-SD1.5 or Lightning-LoRA/Lightning-UNet\citep{lin2024sdxl} for quickly producing 1024$\times$1024 resolution images on HiCo-SDXL.

Multi-concept\citep{kumari2023multi,zhong2024multi,shah2023ziplora} controllable layout generation effectively blend different elements such as characters, style, and objects into a cohesive image. Integrating multi-LoRA models of the same type into the UNet can easily lead to confusion of different conceptual features. This will make it difficult to naturally generate different conceptual elements in the same image. Specifically, training LoRA on the HiCo branch, as shown in Figure \ref{fig:HiCo Lora}, has been shown to significantly enhance performance in scenarios requiring the injection of multiple concepts. For more results, please refer to Appendix \ref{sec:More Generation Results}.

\section{Evaluation}

\subsection{Dataset}
The HiCo model can use various types of grounding data to achieve controllable layout generation across different scenarios.

\textbf{Open-ended Grounded Text2Image Generation}. For training datasets, the fine-grained detailed description data, comprises 1.2 million image-text pairs with regions and descriptions sourced from GRIT-20M\citep{peng2023kosmos}. We performed algorithms and manual cleaning on the raw data, resulting in a dataset with an average of 4.3 objects per image.

\textbf{Closed-set Grounded Text2Image Generation}. For training datasets, the coarse-grained categorical description data, we selecte a subset of approximately images from COCO Stuff\citep{caesar2018coco} based on criteria such as region size, labeled as COCO-75K. This dataset includes 171 categories and an average of 5.5 objects per image.

\textbf{Evaluation Dataset}. The evaluation datasets include two types of data: the coarse-grained COCO-3K and the fine-grained HiCo-7K. We introduce the HiCo-7K benchmark for evaluation of multi-objective controllable layout in natural scenes. The HiCo-7K dataset is derived from GRIT-20M and has undergone iterative cleaning through both algorithmic and manual processes. It consists of 7,000 images, with an average of 3.78 objects per image. We have detailed the construction pipeline of the custom dataset HiCo-7K in Appendix \ref{sec:HiCo-7K Benchmark}.

\subsection{Experimental Setup} \label{Experimental Setup}

We can apply the HiCo architecture to various network structures such as SD1.5, SD2.1, or SDXL\citep{podell2023sdxl} to achieve controllable generation. Specifically, for SD1.5, We utilize the AdamW optimizer with a fixed learning rate of 1e-5 and train the model for 50,000 iterations with a batch size of 256. We train HiCo with 8 A100 GPUs for 3 days. Training HiCo-SDXL requires more iterations and fine-tuning on a smaller set of high-quality data. After training, our HiCo model also supports rapid generation plugins with LoRA, LCM, SDXL-Lightning.

\subsection{Quantitative Results}
\subsubsection{Coarse-Grained Closed-set Text2Img Generation.}

We train and evaluate HiCo model on COCO-stuff datasets to develop its layout-to-image capabilities. We use Frechet Inception Distance(FID)\citep{heusel2017gans} to evaluate the perceptual quality of the generated images. We use YOLO score\citep{li2021image} to evaluate the recognizability of the object in the generated images. YOLO score uses a pretrained YOLOv4\citep{bochkovskiy2020yolov4} model to detect the objects in the generated images.

\begin{figure}[H]
    \vspace{-5mm}
    \centering
    \includegraphics[width=1.0\linewidth]{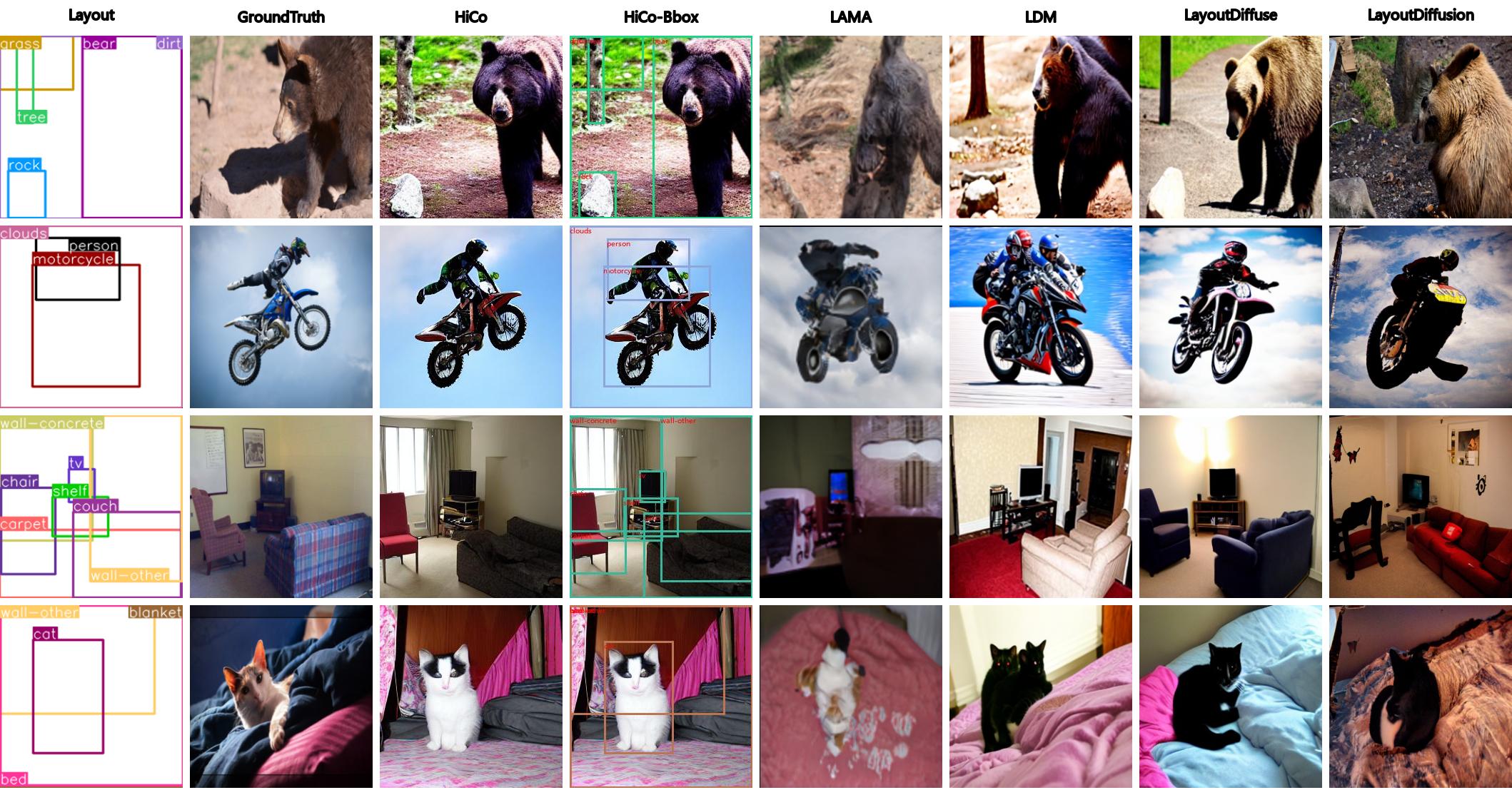}
    \caption{Qualitative comparison of HiCo and other methods on COCO-3K. Compared with other methods, HiCo can generate high-quality images under complex layout conditions. The cases generated by LAMA, LDM and LayoutDiffuse is referenced from LayoutDiffuse.}
    \vspace{-5mm}
    \label{fig:coco_val_figure}
\end{figure}

The HiCo model achieves the best results in image fidelity and spatial semantic dimensions, and the generated images have a more beautiful and reasonable visual effect as shown in Figure \ref{fig:coco_val_figure}. Meanwhile, it can generate images with a resolution of 512 and not just the categories of COCO-3K, which is an ability that other models do not possess. The quantitative comparison is presented in Table \ref{tab:coco_eval_table}.

\begin{table}[H]
    \vspace{-5mm}
    \centering
    \caption{Quantitative comparison of HiCo and other methods on COCO-3K. All generated images are evaluated under 256$\times$256 resolution. ${\dagger}$  indicates that the experimental value is referenced from LayoutDiffuse.}
    \begin{tabular}{cccccc}
        \toprule
        
        \textbf{Methods}	&\textbf{FID}$\downarrow$	&\textbf{AP}$\uparrow$	&\textbf{AP50}$\uparrow$	&\textbf{AP75}$\uparrow$	&\textbf{AR}$\uparrow$\\
        \midrule
        LAMA${\dagger}$\citep{li2021image}	                        &31.12	&20.61	&28.54	&22.69	&26.48\\
        LDM${\dagger}$\citep{rombach2022high}	                    &24.60	&32.13	&54.23	&34.34	&39.86\\
        LayoutDiffuse${\dagger}$\citep{cheng2023layoutdiffuse}	    &20.27	&36.58	&\textbf{59.59}	&38.06	&46.09\\
        LayoutDiffusion\citep{zheng2023layoutdiffusion}	&48.77	&14.20	&24.2	&14.3	&16.2\\
        HiCo(Ours)	                                    &\textbf{20.02}	&\textbf{36.60}	&58.10	&\textbf{39.4}	&\textbf{46.60}\\

        \bottomrule
    \end{tabular}
    \vspace{-5mm}
    \label{tab:coco_eval_table}
\end{table}

\subsubsection{Fine-Grained Open-ended Text2Img Generation.}
\begin{figure}[H]
    \centering
    \includegraphics[width=1.0\linewidth]{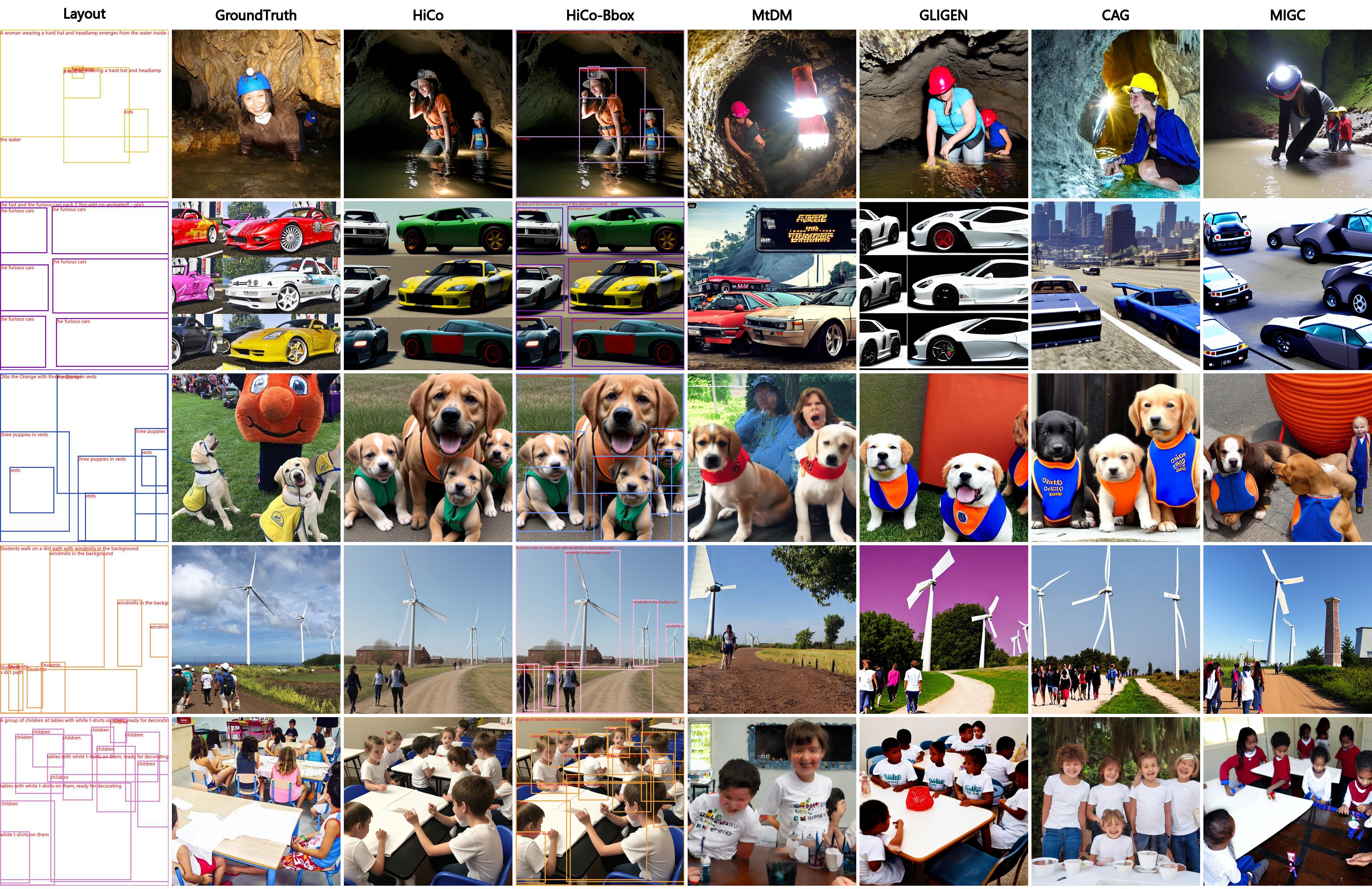}
    \caption{Qualitative comparison with other models on HiCo-7K. Compared with other methods, HiCo can generate high-quality images for both simple and complex layout information.}
    \vspace{-5mm}
    \label{fig:grit_val_figure}
\end{figure}

We train and evaluate HiCo using a high quality data of natural scenes. We use FID, Inception Score (IS)\citep{salimans2016improved}, Learned Perceptual Image Patch Similarity (LPIPS)\citep{zhang2018unreasonable} to evaluate the perceptual quality of the images generated with layout information. The results are presented in Table \ref{tab:grit_eval_table}. Our model can generate high-quality images with rich objects even in complex scenarios, as shown in Figure \ref{fig:grit_val_figure}.

Furthermore, we use Grounding-DINO\citep{liu2023grounding} to detect the instance caption and calculate the maximum IoU between the detection boxes and the ground truth box. If the maximum IoU is higher than the threshod 0.5 and the Local CLIP Score\citep{avrahami2023spatext} of them is higher than 0.2, we mark it as position correctly generated. We use AR, AP, AP50 and AP75 to evaluate the spatial performance. The results are presented in Table \ref{tab:grit_eval_table_spacial}.

\begin{table}[H]
    \vspace{-5mm}
    \centering
    \caption{Quantitative comparison of perception dimension with other models on HiCo-7K. All generated images are evaluated under 512$\times$512 resolution. The results indicate that HiCo has better fidelity and perception.}
    \begin{tabular}{cccc}
        \toprule
        \textbf{Methods}	&\textbf{FID}$\downarrow$	&\textbf{IS}$\uparrow$	&\textbf{LPIPS}$\downarrow$ \\
        \midrule
        MtDM\citep{bar2023multidiffusion}       &24.83      &25.27$\pm$1.54	    &0.76$\pm$0.05  \\
        GLIGEN\citep{li2023gligen}              &19.65	    &26.59$\pm$1.45	    &\textbf{0.73$\pm$0.07}  \\
        CAG\citep{chen2024training}             &19.37	    &26.24$\pm$1.42	    &0.76$\pm$0.06  \\
        MIGC\citep{zhou2024migc}                &26.92	    &22.17$\pm$1.16	    &0.77$\pm$0.07  \\
        InstanceDiff\citep{wang2024instancediffusion}                &16.99	    &26.19$\pm$1.09	    &\textbf{0.73$\pm$0.06}  \\
        HiCo(Ours)                              &\textbf{14.24}	    &\textbf{28.31$\pm$0.79}	    &\textbf{0.73$\pm$0.07}  \\

        \bottomrule
    \end{tabular}
    \vspace{-5mm}
    \label{tab:grit_eval_table}
\end{table}

\begin{table}[H]
    \vspace{-5mm}
    \centering
    \caption{Quantitative comparison of spatial location dimensions on HiCo-7K. Experiments show that HiCo has better positional control and image text consistency.}

    \begin{tabular}{ccccccc}
    \toprule
    \textbf{Methods}	&\textbf{LocalCLIP Score}$\uparrow$	&\textbf{LocalIoU Score}$\uparrow$	&\textbf{AR}$\uparrow$ &\textbf{AP}$\uparrow$ &\textbf{AP50}$\uparrow$ &\textbf{AP75}$\uparrow$\\
    \midrule
    GroundT             &25.07	&66.40	&49.45	&30.13	&42.77	&31.75\\          
    \midrule
    MtDM                &24.81	&25.02	&5.21	&0.59	&2.45	&0.13\\
    GLIGEN              &24.96	&48.54	&28.33	&12.67	&24.15	&12.14\\
    CAG                 &24.54	&24.43	&4.62	&0.47	&2.01	&0.21\\
    MIGC                &24.94	&48.03	&25.8	&10.18	&23.42	&7.56\\
    InstanceDiff        &25.09	&51.58	&33.92	&16.76	&27.45	&17.14\\
    HiCo(Ours)          &\textbf{25.17}	&\textbf{59.98}	&\textbf{41.21}	&\textbf{21.53}	&\textbf{33.22}	&\textbf{22.66}\\
    
    \bottomrule
    \end{tabular}
    \vspace{-5mm}
    \label{tab:grit_eval_table_spacial}
\end{table}

\textbf{Zero-shot Evaluation}. We further evaluate the zero-shot performance of HiCo trained in natural scenes on COCO-3K, detailed results are shown in table \ref{tab:coco_eval_table_spacial}. The preferences of HiCo controllability is not the best on COCO-3K. The reason for this problem is that our model was trained on a 1.2M fine-grained long caption, which belongs to out of distribution data for COCO data.

\begin{table}[H]
    \vspace{-5mm}
    \centering
    \caption{Quantitative comparison of spatial location dimensions of zero-shot capability on COCO-3K.}
    \begin{tabular}{ccccccc}
    \toprule
    \textbf{Methods}	&\textbf{LocalCLIP Score}$\uparrow$	&\textbf{LocalIoU Score}$\uparrow$	&\textbf{AR}$\uparrow$ &\textbf{AP}$\uparrow$ &\textbf{AP50}$\uparrow$ &\textbf{AP75}$\uparrow$\\
    \midrule
    GLIGEN              &25.83	&73.17	&53.39	&36.87 &65.35 &38.09 \\
    MIGC                &25.94	&75.06	&57.55	&40.82 &70.57 &42.21 \\
    InstanceDiff        &\textbf{26.60}	&\textbf{85.91}	&\textbf{78.10}	&\textbf{65.28} &\textbf{80.56} &\textbf{71.93} \\
    HiCo-Real(Ours)     &\underline{26.27}	&\underline{79.57}	&\underline{68.40}	&\underline{52.47} &\underline{78.61} &\underline{57.05} \\
    
    \bottomrule
    \end{tabular}
    \vspace{-5mm}
    \label{tab:coco_eval_table_spacial}
\end{table}

\subsection{Human Evaluation}

We use a multi-round and multi-participant cross-evaluation approach to assess human preferences, focusing on aspects such as object quantity, spacial location, and global image quality. Details on the experimental setup can be found in supplementary material of Appendix \ref{sec:Manual Evaluation Criteria}.

\textbf{Spatial Location \& Semantics}.Target quantity dimension assesses whether the number of objects in the generated image aligns with the preset value. The semantics and position dimension examine the relevance of the objects to the textual description and their regional placement within the image.

\textbf{Global Image Quality}. The image global quality dimension includes five sub-dimensions: relativity, clarity, rationality, aesthetics and risk. 

\begin{table}[H]
    \vspace{-5mm}
    \centering
    \caption{The human evaluation results encompass two dimensions: spatial semantic location and global image quality. The numerical range is from 0 to 100, with a higher score indicating better performance. It should be noted that the risk dimension is the proportion of generated risk images, the lower the better.}
    \begin{tabular}{p{.10\textwidth}m{.07\textwidth}m{.06\textwidth}m{.07\textwidth}m{.07\textwidth}m{.05\textwidth}m{.08\textwidth}m{.06\textwidth}m{.03\textwidth}m{.06\textwidth}}
        \toprule
        \multirow{2}{*}{Method}  & \multicolumn{3}{c}{Spetial Location\& Semantics} & \multicolumn{5}{c}{Global Image Quality} & \multirow{2}{*}{Overall} \\ \cline{2-9}
        & quantity & semantics & position  & relativity  & clarity  & rationality  & aesthete  & risk$\downarrow$  \\ 
        \midrule
        GroundT        & 100   &100    &100    &93.36  &90.69  &97.34  &92.56 &0.00  &95.62 \\ 
        SD-Real       & 80.56	& -	    & -	    &67.41	&83.58  &60.62	&63.67 &2.00  & -    \\
        \toprule
        MtDM           & 54.96	&33.26  &22.04  &48.08  &59.04  &40.49	&40.91 &9.00  &39.41 \\
        GLIGEN         & 80.44  &73.56  &72.78  &56.78  &61.67  &53.67  &51.89 &4.44  &67.76 \\
        CAG            & 50.89	&32.11	&37.89	&48.67	&55.11	&49.00	&44.89 &7.33  &43.94 \\
        MIGC           & 69.89	&62.44	&62.67	&49.22	&57.44	&47.00	&46.67 &6.11  &59.03 \\
        InstanceDiff   & 88.33	&80.00	&66.78	&61.56	&62.33	&56.00	&55.33 &2.50   &70.54 \\
        HiCo           & \textbf{89.44}	&\textbf{86.11}	&\textbf{82.22}	&\textbf{63.67}	&\textbf{82.33}	&\textbf{58.33}	&\textbf{60.89} &\textbf{2.00}  &\textbf{78.08} \\
        \bottomrule
        % \underline{}  \textbf{}
    \end{tabular}
    \vspace{-5mm}
    \label{tab:human evaluation results}
\end{table}

Table \ref{tab:human evaluation results} demonstrates the human evaluation results conducted on 300 controllable layout images. The results indicate that, in terms of spatial position and semantic dimension, our approach outperforms other models. Moreover, it achieves near-par performance to the RealisticVisionV51 model(SD-Real) in the fine-grained dimension of global image quality, suggesting that despite the enhanced controllability, the generative capabilities of our model remain robust and effective.

\subsection{Ablation Studies} \label{Ablation Studies}
The ablation focuses on UNetGlobalCaption(\textbf{UGC}), GlobalBackgroundBranch(\textbf{GBB}) and FuseNet(\textbf{FN}). Furthermore, FuseNet is a non-parametric network that includes the following types, such as (1)Summation. (2)Average. (3)Mask. Experiments are performed on HiCo-7K using the HiCo-base model with the same training amount. The results are presented in Table \ref{tab:grit_eval_table_ablation}.

\begin{table}[H]
    \vspace{-5mm}
    \centering
    \caption{The results of ablation studies on HiCo-7K of UGC, GBB and FN.}
    \begin{tabular}{ccc|ccccc}
        \toprule
        \textbf{UGC}    &\textbf{GBB}	&\textbf{FN}    &\textbf{FID}$\downarrow$   &\textbf{AR}$\uparrow$    &\textbf{AP}$\uparrow$ &\textbf{LocalCLIP Score}$\uparrow$   &\textbf{LocalIoU Score}$\uparrow$ \\
        \midrule
        $\times$        &$\checkmark$  &sum  &20.81  &33.12  &16.54  &25.18  &52.25  \\
        $\times$        &$\checkmark$  &avg  &18.78  &36.3   &18.58  &25.02  &56.67  \\
        $\times$        &$\checkmark$  &mask &\textbf{15.26}  &\textbf{39.51}  &\textbf{20.02}  &25.22  &\textbf{59.07}  \\
        $\checkmark$    &$\checkmark$  &mask &16.85  &38.19  &19.05  &25.23  &57.03  \\
        $\checkmark$    &$\times$      &mask &21.82  &30.07  &12.07  &\textbf{25.24}  &49.64  \\
        \bottomrule
    \end{tabular}
    \vspace{-5mm}
    \label{tab:grit_eval_table_ablation}
\end{table}

\subsection{Inference Performance}

\begin{figure}[H]
    \vspace{-5mm}
    \centering
    \includegraphics[width=0.8\linewidth]{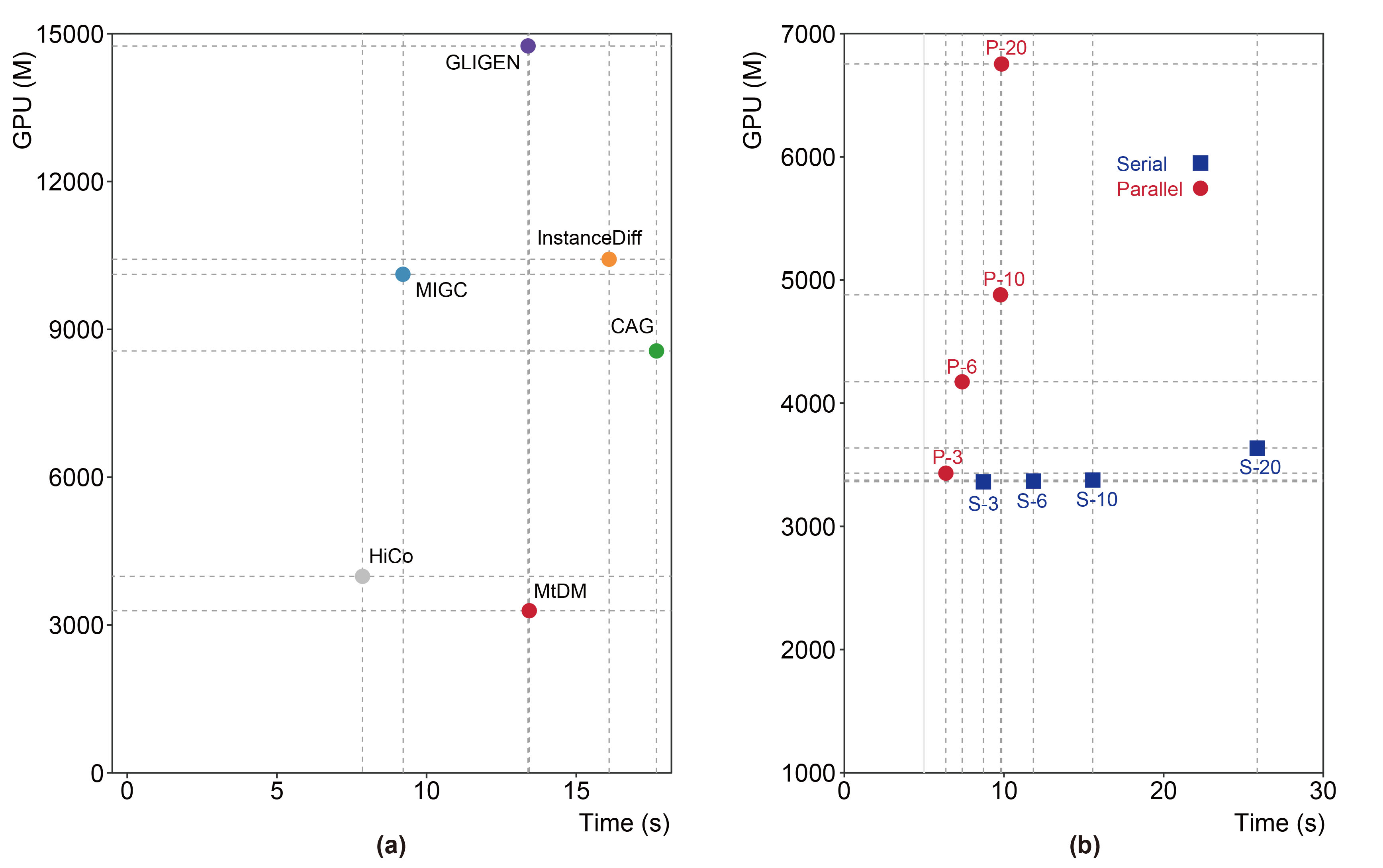}
    \caption{Quantitative comparison of inference performance. (a) Comparison between different methods on HiCo-7K. (b) Comparison between different object quantities on HiCo.}
    \vspace{-5mm}
    \label{fig:HiCo Performance}
\end{figure}

For inference run time and memory usage, we conducted two additional comparisons. The first comparison is horizontal, among 6 different models including: GLIGEN, InstanceDiff, MIGC, CAG, MtDM as well as our HiCo. Specifically, we evaluated the inference time and GPU memory usage for directly generating 512$\times$512 resolution images on the HiCo-7K using a 24GB VRAM 3090 GPU, results in Figure \ref{fig:HiCo Performance}-(a) show that HiCo has the shortest inference time and the 2nd lowest GPU memory footprint.

The multi-branch of HiCo has two inference modes: "parallel mode" and "serial mode". In order to verify the performance advantages of HiCo when the number of objects increases, the second comparison is vertical: we assessed the inference time and GPU memory usage for generating 512$\times$512 resolution images on the HiCo-7K with different number of objects. Since each object is processed by a separate branch in HiCo, the inference can be accelerated by inferring all the branches in one batch, in "parallel mode", which as shown in Figure \ref{fig:HiCo Performance}-(b), is much faster than the "serial mode", inferring all the branches one by one in serial.

\section{Conclusion} \label{Conclusion}
HiCo is a controllable layout generation model based on diffusion model, guided by multiple branch structures. This approach allows users to specify the location and detailed textual descriptions of target regions while maintaining the rationality and controllability of the generated content. Through training and testing on data with varying degrees of granularity in natural scenarios, as well as conducting algorithm metric evaluation and subjective human assessment, the superiority of this method is demonstrated. However, there is still potential for further improvement, especially in the areas of image content editing and integrating multiple style concepts. By combining current controllable generation capabilities can boost the overall playability of AI-generated artwork.

\textbf{Limitation}. The complex interactions and occlusion order of overlapping areas are significant challenges for HiCo model and even the field of layout-to-image generation. HiCo achieves hierarchical generation by decoupling each object’s position and appearance information into different branch, meanwhile controlling their overall interactions through a background branch with global prompt and the Fuse Net. HiCo is capable of handling complex interactions in overlapping regions by FuseNet. The occlusion order of overlapping objects is also specified via the global prompt by text description. But since there lacks corresponding occlusion order train data, the success rate is far from optimal. For current version of HiCo, there indeed lacks of more explicit mechanism for occlusion order controlling. For more results, please refer to Appendix \ref{sec:Limitaion Discussion}. We also found that the HiCo model still does not handle the generation of complex layouts for multiple concepts of LoRA very well. We intend to conduct further research to explore solutions to these issues in the future. 

\textbf{Social impact aspect}. Our model is designed to assist users in generating creative images with controllable layouts. However, there is a risk of misuse of our method to generate inappropriate or sensitive content. Therefore, we believe that regulating the application of such models and developing risk detection tools are crucial. This will facilitate the progress of AI technology for the benefit of humanity.

\clearpage

{
\small
\bibliography{reference}
}

\medskip

%%%%%%%%%%%%%%%%%%%%%%%%%%%%%%%%%%%%%%%%%%%%%%%%%%%%%%%%%%%%

\appendix

%%%%%%%%%%%%%%%%%%%%%%%%%%%%%%%%%%%%%%%%%%%%%%%%%%%%%%%%%%%%

\newpage
\section{Appendix / supplemental material}
\section*{Overview}
In this supplemental material, we provide the following items:
\begin{itemize}[leftmargin=*]
    \item Training and Inference Strategies.
    \item HiCo-7K Benchmark.
    \item Limitaion Discussion.
    \item Manual Evaluation Criteria.
    \item More results on HiCo, including different layout quantities, different base models, and different resolutions.
\end{itemize}

\subsection{Training and Inference Strategies} \label{sec:Training and Inference Strategies}

During the training stage, we only optimize the parameters of the HiCo Net, while keeping the SD base model parameters fixed. The Fuse Net can use either non-parametric methods, like summation, averaging, mask or a parametric method, like simple MLP structure.

During the inference stage, the structure of the Fuse Net can be reasonably selected according to the size and importance of the foreground and background regions to achieve controlled generation effects in different scenarios. Additionally, LoRA network parameters can be added to the HiCo Net to fine-tune and learn new tasks, such as personalization, multi-language controllable generation and other tasks.

\subsection{HiCo-7K Benchmark} \label{sec:HiCo-7K Benchmark}

\begin{figure}[H]
    \centering
    \includegraphics[width=0.8\linewidth]{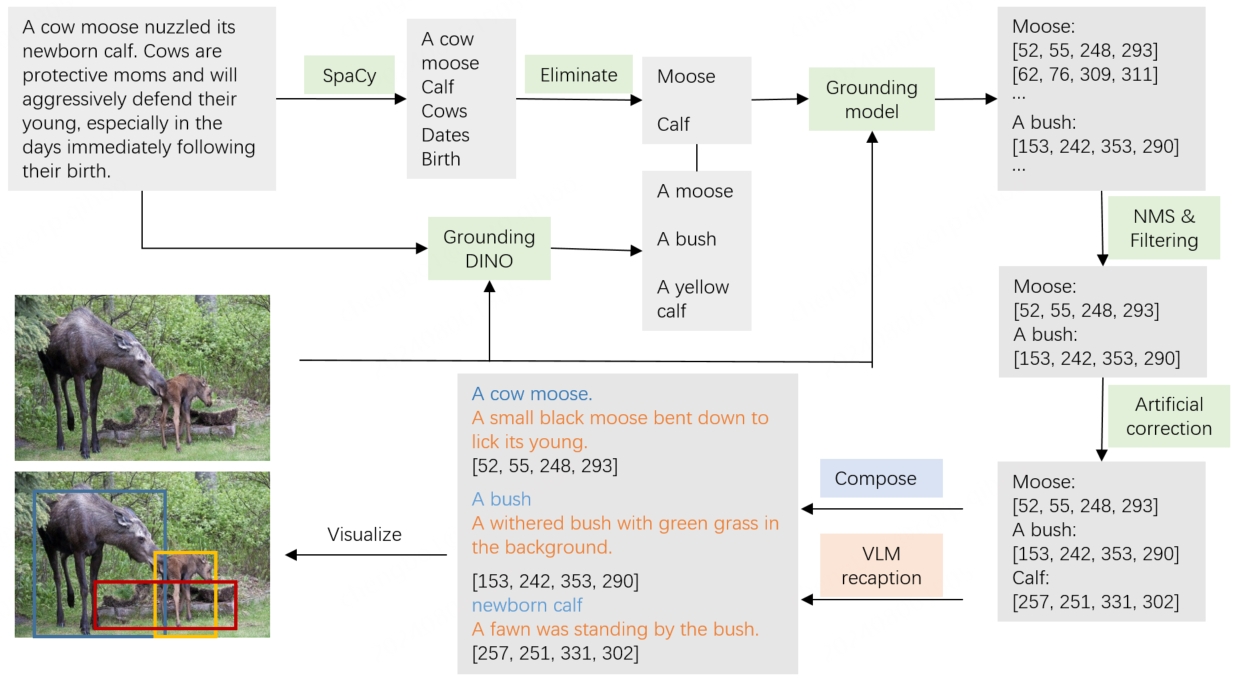}
    \caption{The pipeline of constructing HiCo-7K with grounded image caption pairs.}
    \label{fig:HiCo-7K benchmark}
    \vspace{-5mm}
\end{figure}

We have detailed the construction pipeline of the custom dataset HiCo-7K in Figure \ref{fig:HiCo-7K benchmark}. We found that GRIT-20M has some issues, such as a low labeling rate for objects with the same description and target descriptions being derived solely from the original captions. Compared to GRIT-20M, the pipeline of the HiCo-7K is as follows.
\begin{enumerate}[leftmargin=*]
    \item Extracting noun phrase. We use spaCy to extract nouns from captions and the LLM VQA model to remove abstract noun phrases. Meanwhile, we use the GroundingDINO model to extract richer phrase expressions.
    \item Grounding noun phrase. We use the GroundingDINO model to obtain the bboxes. After that, we use NMS and CLIP algorithms to clean the bboxes.
    \item Artificial correction. To address the issue of algorithmic missed detections for multiple objects with the same description in an image, artificial correction is employed to further enhance the labeling rate of similar objects.
    \item Multi-captions with bounding box. We expand the basic text from the original captions and use GPT-4 to re-caption the target regions. The dataset of HiCo-7K contains 7000 expression-bounding-box pairs with referring-expressions and GPT-generated-expressions.
\end{enumerate}

\subsection{Limitaion Discussion} \label{sec:Limitaion Discussion}

\begin{figure}[H]
    \centering
    \includegraphics[width=0.8\linewidth]{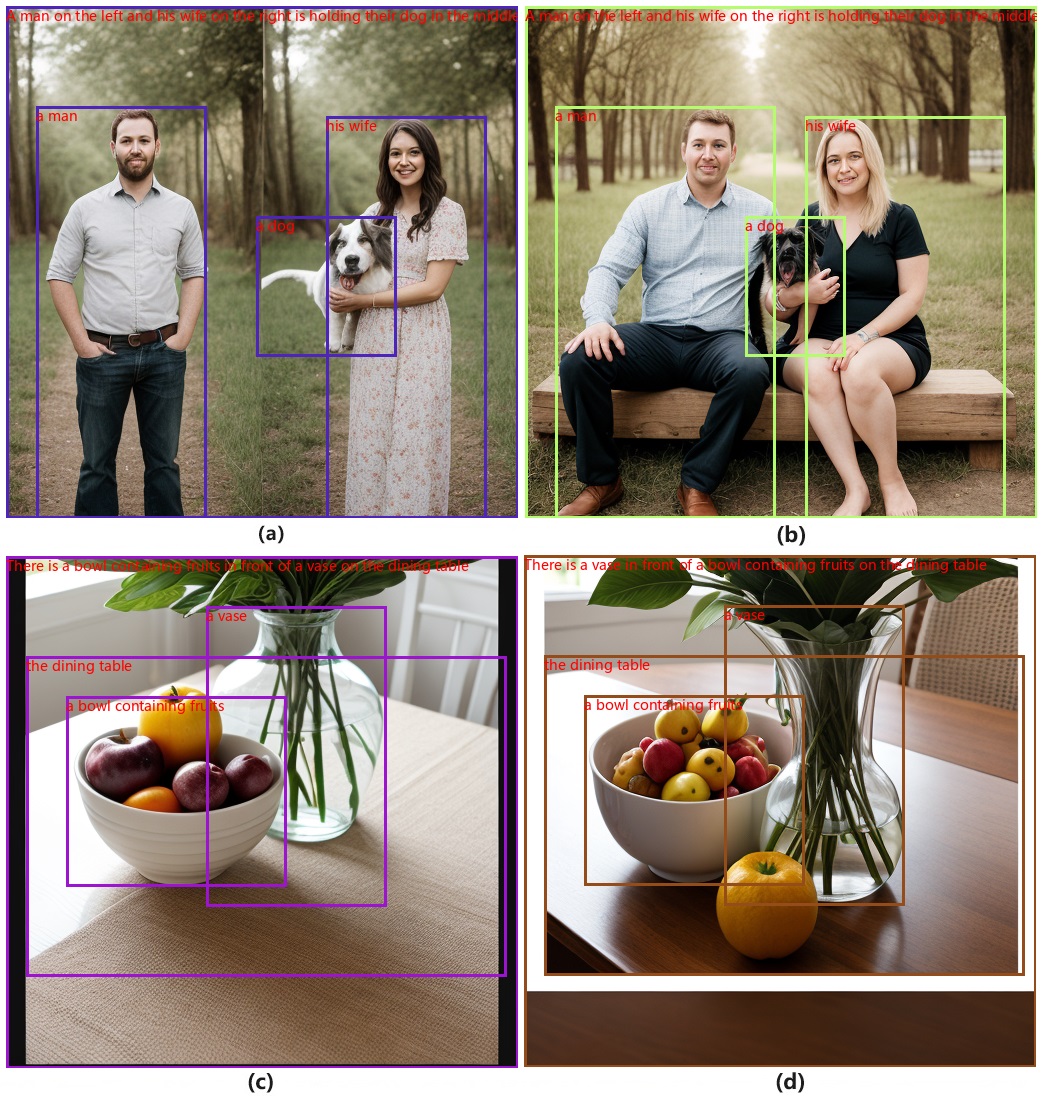}
    \caption{The image generated in overlapping and interactive scenarios.The generated caption is displayed in the image.}
    \label{fig:HiCo_Overlap_Analsys}
    \vspace{-5mm}
\end{figure}

HiCo achieves hierarchical generation by decoupling each object’s position and appearance information into different branch, meanwhile controlling their overall interactions through a background branch with global prompt and the Fuse Net. The Fuse Net combines features from foreground and background regions, as well as intermediate features from side branches, then integrates them during the UNet upsampling stage. As illustrated in Figure \ref{fig:HiCo_Overlap_Analsys}-(a),Figure \ref{fig:HiCo_Overlap_Analsys}-(b), HiCo is capable of handling complex interactions in overlapping regions without any difficult.The occlusion order of overlapping objects is also specified via the global prompt by text description, for example “bowl in front of vase”, as illustrated in Figure\ref{fig:HiCo_Overlap_Analsys}-(c),Figure\ref{fig:HiCo_Overlap_Analsys}-(d). But since there lacks corresponding occlusion order train data, the success rate is far from optimal. For current version of HiCo, there indeed lacks of more explicit mechanism for occlusion order controlling.

We recognize this problem as our future work. Actually we’re already working on the occlusion order data curation, which is a quite challenging task as it requires reliable depth estimation besides the object detection bounding boxes.

\subsection{Manual Evaluation Criteria} \label{sec:Manual Evaluation Criteria}
\subsection*{Evaluation Dimensions}
We designed the evaluation dimensions by reviewing existing literature and soliciting opinions from professional designers and a large number of users. Specifically, we divided the evaluation dimensions into two categories: local and global. Local dimensions include spatial location and semantics, such as the quantities, semantics, and positions of the object bounding boxes. Global dimensions include global image quality, such as relativity, clarity, rationality, aesthetics, risk, and overall score.

\subsection*{Dimension Definitions}
Spatial Location\& Semantics. Local dimensions include three sub-dimensions: quantity indicates the number of generated objects; semantics measures the consistency between the objects within the bounding box and the textual description of that region; position measures the deviation between generated objects and ground truth by calculating the IoU of bounding boxes.

Global Image Quality. The image global quality dimension includes five sub-dimensions: relativity is primarily used to evaluate the understanding and representation of text by image content; clarity is a commonly used metric for assessing image quality; rationality  is used to describe the distortion, deformation, and disarray of image content; aesthetics encompasses the overall assessment of the visual appeal of the generated image, incorporating factors such as detailed depiction, color usage, creativity, and other relatively subjective judgment elements; risk evaluates elements related to nudity, violence, terror, and other sensitive content within the image.

For risk we assess the presence of such elements and represent it using binary values of 0 or 1. For the other dimensions, we categorize them into four levels ranging from 0 to 3 and then normalize them to 0-100. We use overall score to represent the comprehensive assessment of the image.

\subsection*{Evaluation Execution}
The evaluation team consists of professional evaluators. They have rich professional knowledge and evaluation experience, allowing them to accurately execute the evaluation tasks based on the given dimensions.

Specifically, our evaluation score is computed with the following steps:

\begin{enumerate}[leftmargin=*]
    \item According to the rules provided, evaluators rate each image on each dimension. Risk is scored as either 0 or 1, while the other dimensions range from 0 to 3.
    \item Calculating the overall rate for each image: We calculate the overall rate by summing the weighted scores of each dimension. For the sub-dimensions of the local dimension, the weight is 0.2, while for the sub-dimensions of the global dimension, the weight is 0.1. The final score for overall rate is calculated as the weighted sum of the seven dimensions multiplied by the score of risk (which is either 0 or 1).
    \item We calculate the mean of each dimension across the entire evaluation set, excluding the risk dimension, and mapped the scores to a scale of 0 to 100. For risk, we calculate the coverage rate of this dimension.
\end{enumerate}

\subsection{More Generation Results} \label{sec:More Generation Results}

\begin{figure}[H]
    \centering
    \includegraphics[width=1.0\linewidth]{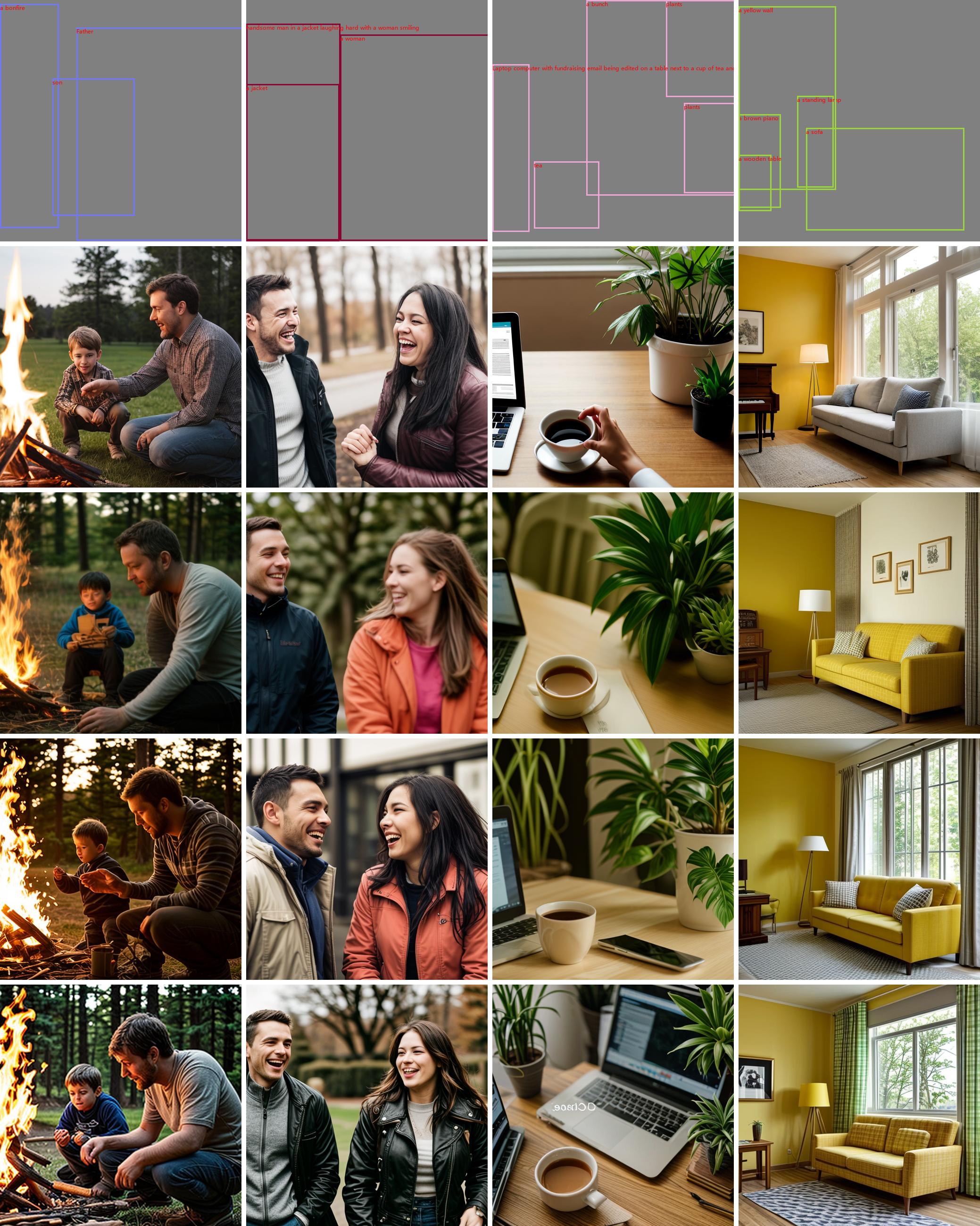}
    \caption{Qualitative experiments. Fast generation of complex layout information with HiCo-LCM / HiCo-Lightning.}
    \label{fig:grit_a3_3}
\end{figure}

Figure \ref{fig:grit_a3_3} shows more results generated by HiCo using fast generated plug-in LoRA. The four columns from left to right represent the generated results of 50-steps, 4-steps, 6-steps, and 8-steps, respectively.

\begin{figure*}[t!]
    % \vspace{-100mm}
    \centering

    \subfigure{
        \begin{minipage}[t]{0.48\textwidth}
            \centering
            \includegraphics[width=1\textwidth]{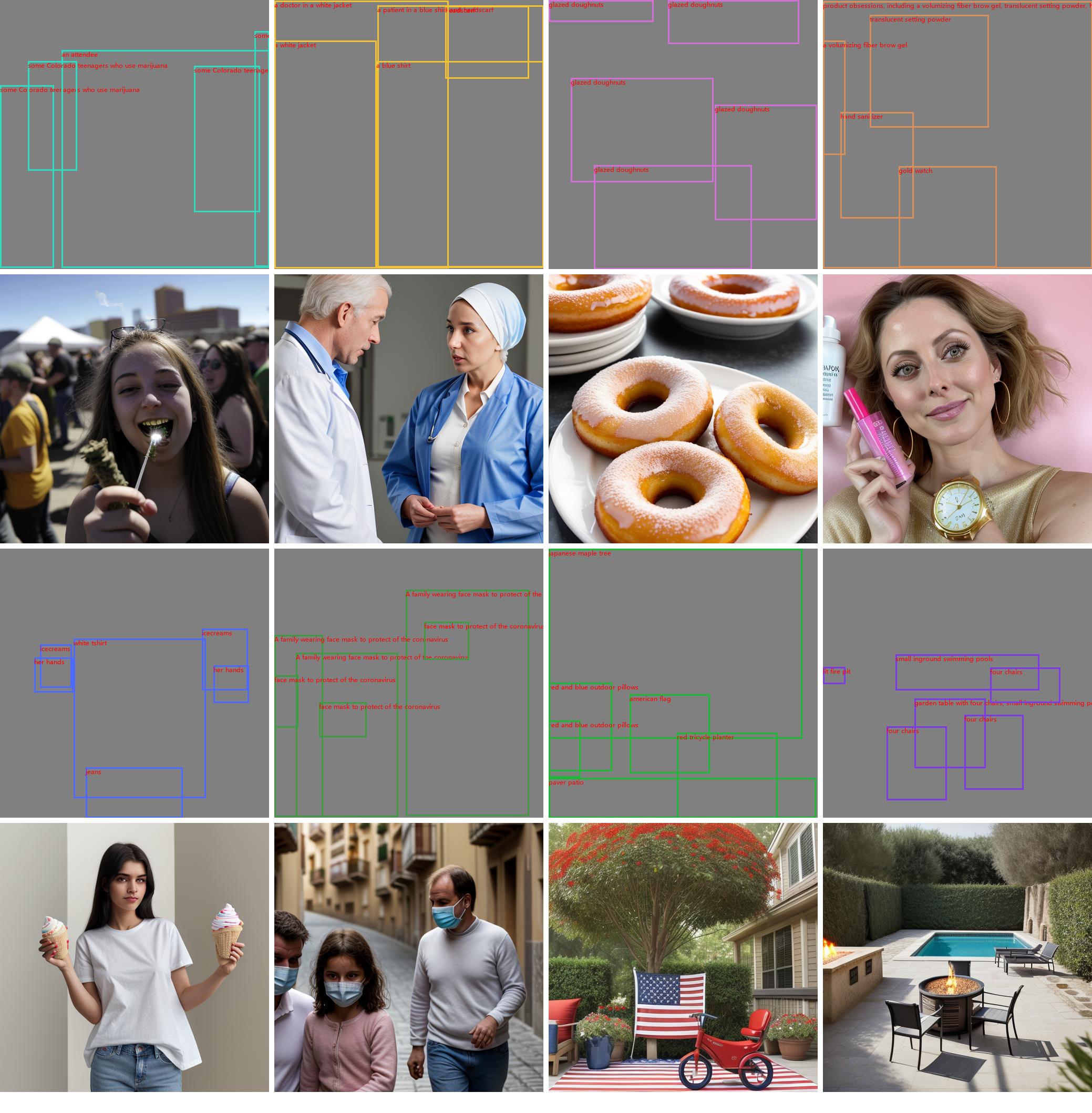}            
            % \vspace{-0.6cm}
            \caption*{ (a) Generation of complex layout information of HiCo-SD1.5.}
            \label{fig:grit_a3_1}
        \end{minipage}
    }
	\subfigure{
        \begin{minipage}[t]{0.48\textwidth}
            \centering
            \includegraphics[width=1\textwidth]{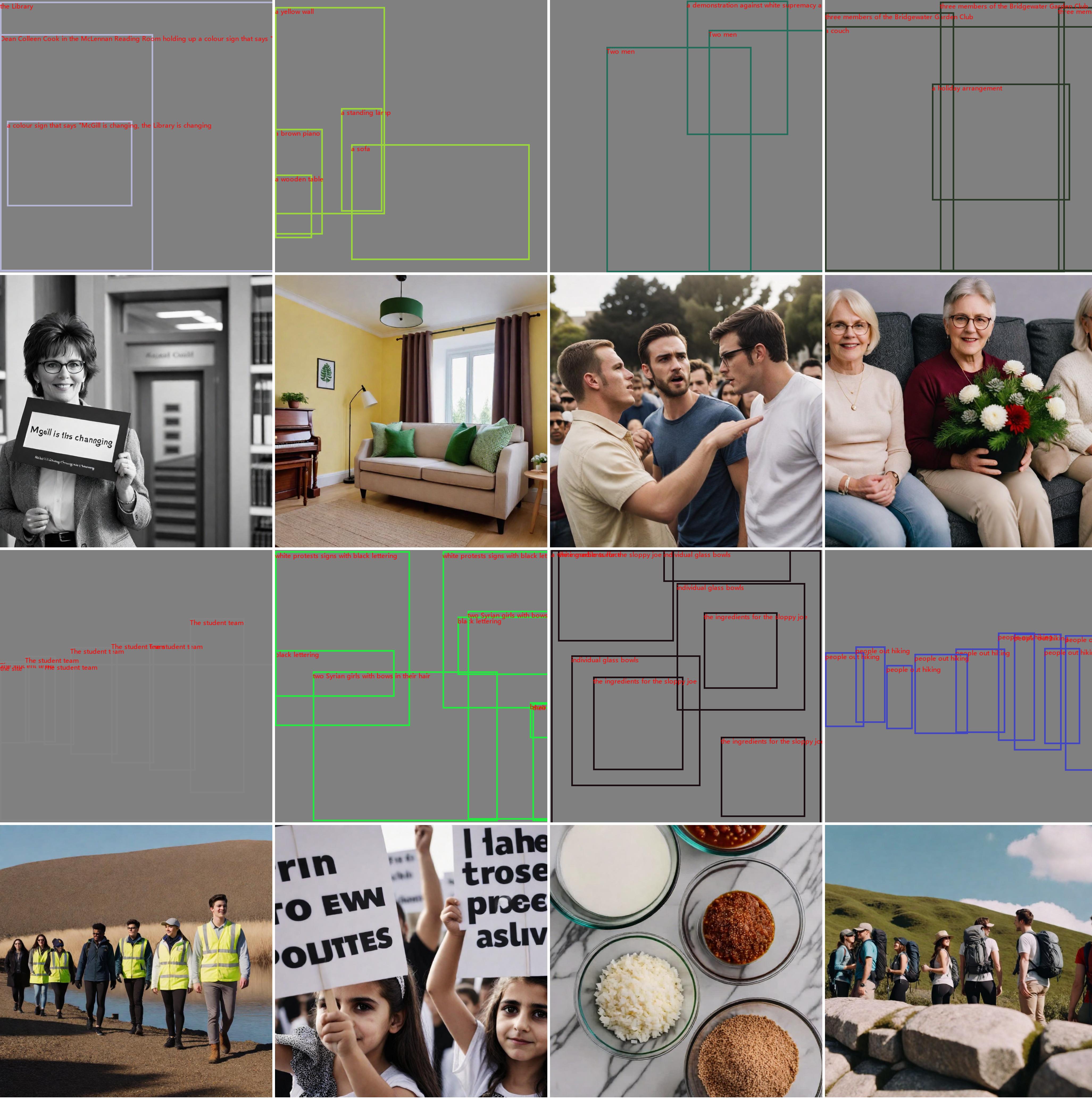}            
            % \vspace{-0.6cm}
            \caption*{ (b) Generation of complex layout information of HiCo-SDXL.}
            \label{fig:grit_a3_2}
        \end{minipage}
    }
    \subfigure{
        \begin{minipage}[t]{0.48\textwidth}
            \centering
            \includegraphics[width=1\textwidth]{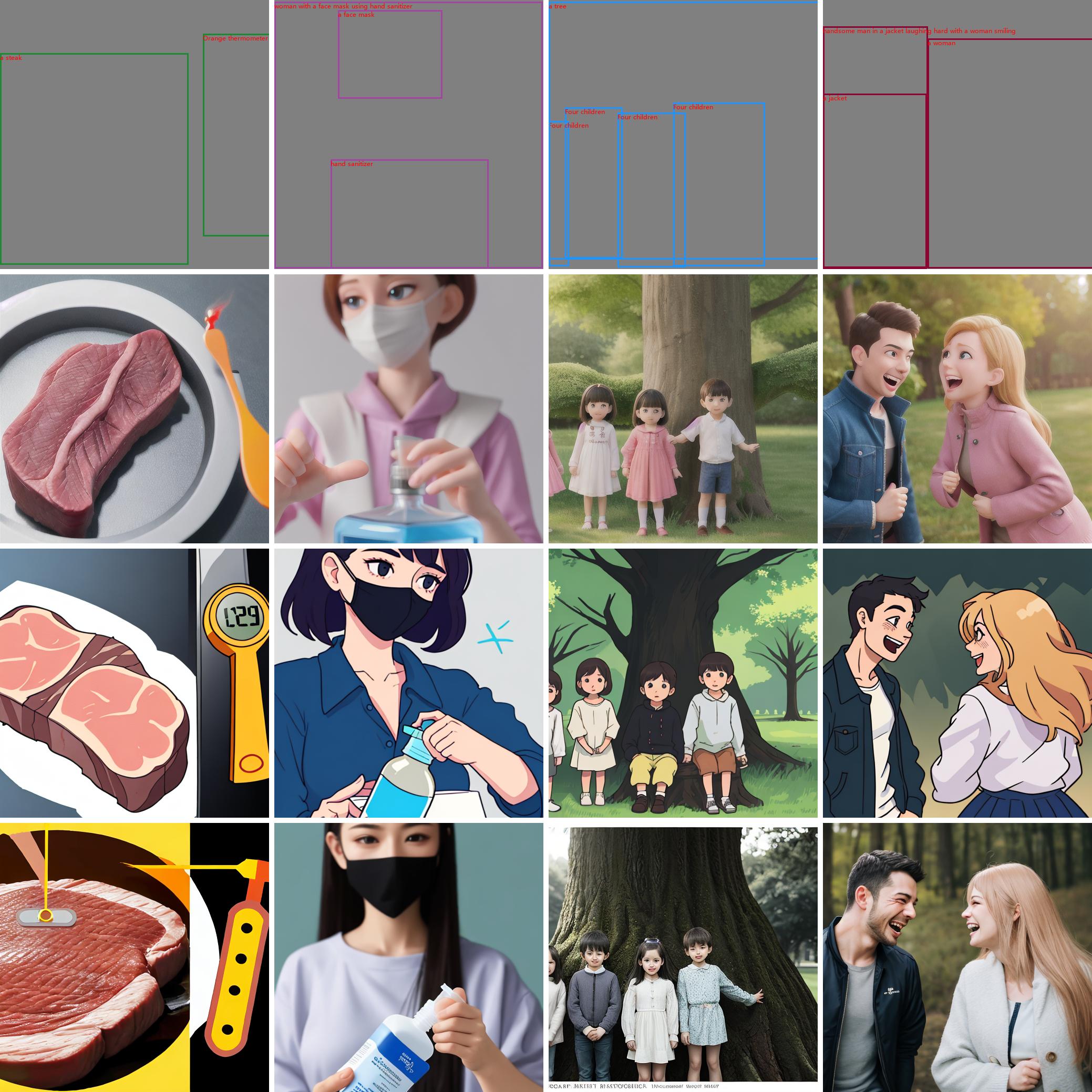}            
            % \vspace{-0.6cm}
            \caption*{ (c) Generation of complex layout information of HiCo with different backbones.}
            \label{fig:grit_a3_4}
        \end{minipage}
    }
	\subfigure{
        \begin{minipage}[t]{0.48\textwidth}
            \centering
            \includegraphics[width=1\textwidth]{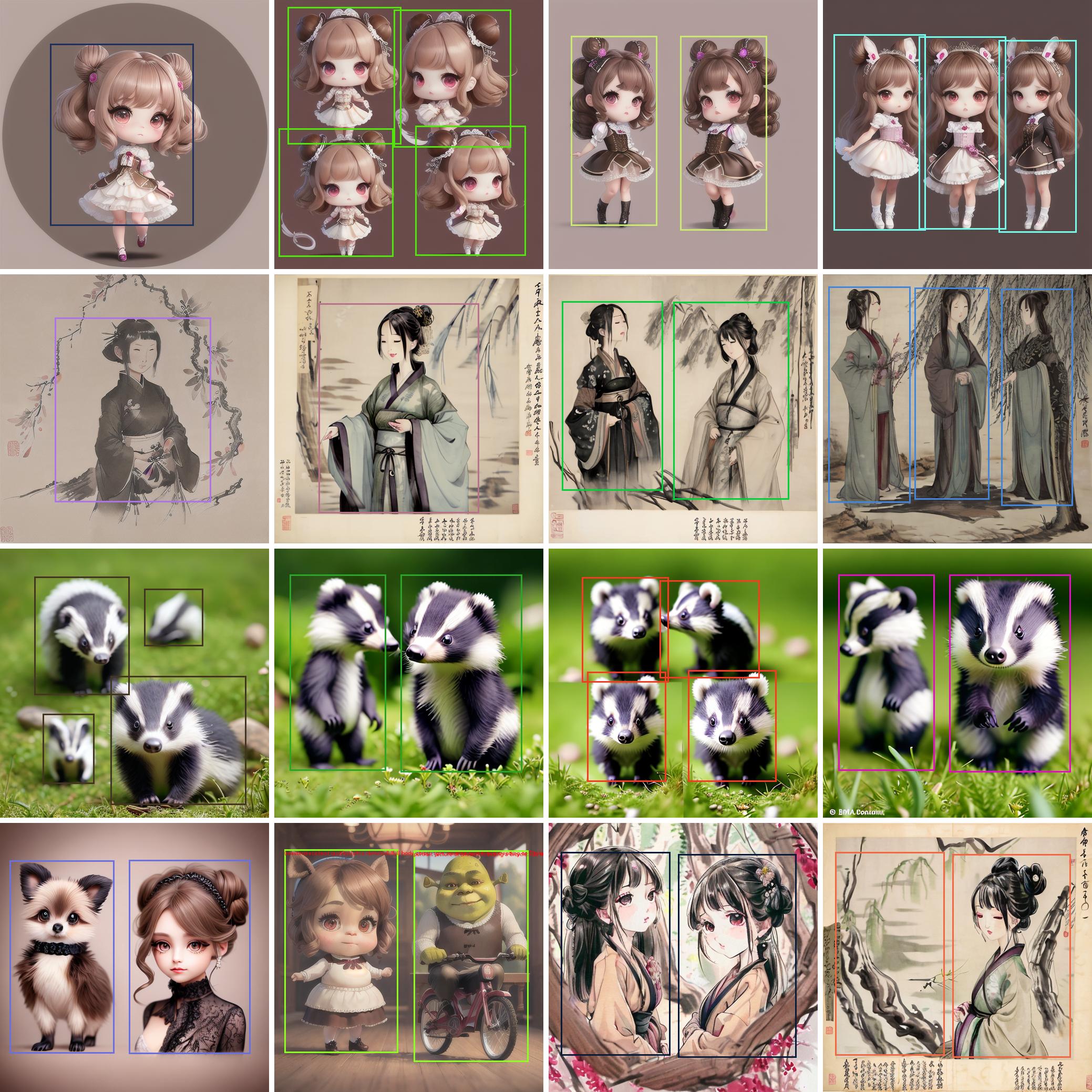}            
            % \vspace{-0.6cm}
            \caption*{ (d) Generation of concepts by HiCo with multi LoRA.}
            \label{fig:grit_a3_5}
        \end{minipage}
    }

    \vspace{-3mm}
    % 添加题注，即对这个图片的说明
	\caption{Qualitative experiments generated by HiCo model to show the layout controllability.}
	\vspace{-3mm}
    \label{fig:appd_base_result}
\end{figure*}

Figure \ref{fig:appd_base_result}-(a) shows more results generated by HiCo-SD1.5 in HiCo-7K. Figure \ref{fig:appd_base_result}-(b) shows more results generated by HiCo-SDXL in HiCo-7K. The number of object layouts ranges from 3 to over 10. Despite complex layouts and rich descriptions, HiCo reliably ensures that each object is generated in the correct position with the right description.

Figure \ref{fig:appd_base_result}-(c) shows more results generated by HiCo using different checkpoint from open source community. The results from the second to fourth lines are generated by the following three models, namely disneyPixarCartoon, flat2DAnimerge and dreamshaper.

Figure~\ref{fig:appd_base_result}-(d)  show the generation of different layout information with LoRA. Rows 1-3  demonstrate that HiCo can effectively generate complex layouts for a single concept with single LoRA. Row 4 shows the generation of complex layouts for multiple concepts with multi LoRAs using the HiCo model. 

Through further experiments, we found that the HiCo model still does not handle the generation of complex layouts for multiple concepts very well. More effective methods need to be explored to address this issue.

% \begin{figure*}[!htbp]

% \begin{figure}[H]
%     \centering
%     \includegraphics[width=1.0\linewidth]{images/appd-A3-1-1.jpg}
%     \caption{Qualitative experiments. Generation of complex layout information of HiCo-SD1.5.}
%     \label{grit_a3_1}
% \end{figure}

% \begin{figure}[H]
%     \centering
%     \includegraphics[width=1.0\linewidth]{images/appd-A3-2.jpg}
%     \caption{Qualitative experiments. Generation of complex layout information of HiCo-SDXL.}
%     \label{grit_a3_2}
% \end{figure}

% \begin{figure}[H]
%     \centering
%     \includegraphics[width=1.0\linewidth]{images/appd-A3-4.jpg}
%     \caption{Qualitative experiments. Generation of complex layout information of HiCo with different backbones.}
%     \label{grit_a3_4}
% \end{figure}

% \begin{figure}[H]
%     \centering
%     \includegraphics[width=1.0\linewidth]{images/appd-A3-5.jpg}
%     \caption{Qualitative experiments. Generation of concepts by HiCo with multi LoRA.}
%     \label{grit_a3_5}
% \end{figure}

%%%%%%%%%%%%%%%%%%%%%%%%%%%%%%%%%%%%%%%%%%%%%%%%%%%%%%%%%%%%
\clearpage 

\end{document}